\documentclass[10pt,twocolumn,letterpaper]{article}

\usepackage[pagenumbers]{wacv} 

\definecolor{wacvblue}{rgb}{0.21,0.49,0.74}
\usepackage[pagebackref,breaklinks,colorlinks,allcolors=wacvblue]{hyperref}

\usepackage{graphicx}
\usepackage{amsmath}
\usepackage{amssymb}
\usepackage{booktabs}
\usepackage{multirow}
\usepackage{tabularx,verbatim}
\usepackage{xspace}
\usepackage{algorithm}
\usepackage{algorithmic}
\usepackage{setspace}
\usepackage{comment}
\usepackage{bbm}
\usepackage{enumitem}
\usepackage{colortbl}
\usepackage{color, xcolor} 
\usepackage[T1]{fontenc}
\usepackage{bbm}
\usepackage{dsfont}
\usepackage{bbding}

\def\wacvPaperID{1195} 
\def\confName{WACV}
\def\confYear{2027}

\title{See and Detect: Image-Driven Open-Vocabulary Multi-View 3D Object Detection}

\author{
    Yating Xu$^{1}$ \quad
    Na Zhao$^{2\dagger}$ \quad
    Chenhan Jiang$^{3}$ \quad
    Qingwen Zhang$^{4}$ \quad
    Jiankang Deng$^{1}$ \quad
    Jifei Song$^{1}$ \\[2pt]
    {\small
        $^{1}$MVP Lab \quad
        $^{2}$Singapore University of Technology and Design
    } \\
    {\small
        $^{3}$Hong Kong University of Science and Technology \quad
        $^{4}$KTH Royal Institute of Technology
    }\\[2pt]
    {\small
        $^{\dagger}$Corresponding author \quad
        Email (Yating Xu): \texttt{xuyt98@gmail.com}
    }
}

\begin{document}
\maketitle
\begin{abstract}
Recent open-vocabulary multi-view indoor 3D detection methods typically follow a point-cloud-driven pipeline, where pseudo 3D boxes are generated from reconstructed or estimated geometry and semantic labels are inferred from projected regions. 
However, such a design makes box generation dependent on the completeness of point clouds and restricts semantic learning to local appearance, causing inferior 3D object localization and open-vocabulary classification. 
In this paper, we present See and Detect, an image-driven framework that uses multi-view images as the primary evidence for 3D box and semantic label generation. 
For box generation, we first discover object candidates from individual images, enabling objects missing in point clouds to be recovered from visual evidence. However, single-view predictions are often noisy and redundant. To address this, we propose multi-view consensus box consolidation to fuse spatially consistent predictions across views, and box quality assessment to reduce the impact of boxes with partial object coverage. 
For semantic label generation, we introduce context-aware semantic learning, where semantic labels are queried from full image context, enabling more accurate semantic prediction.
Extensive experiments on ScanNet and ARKitScenes demonstrate that our method achieves state-of-the-art open-vocabulary multi-view 3D detection performance while running over 500$\times$ faster than the previous state-of-the-art method.
\end{abstract}
    
\section{Introduction}
\label{sec:intro}

Open-vocabulary 3D object detection \cite{cao2023coda,wang2024ov,lu2023open,jiao2024unlocking} is a fundamental task for open-world 3D scene understanding, aiming to localize and classify arbitrary object categories in 3D space. It is essential for embodied applications such as navigation, manipulation, and human--robot interaction. 
Although point-cloud-based methods \cite{cao2023coda, lu2023open} have achieved impressive progress, they rely on depth sensors or LiDAR during deployment to capture geometric inputs, which are expensive, power-intensive, and often unavailable on consumer-grade platforms. 
Recently, detecting 3D objects from posed multi-view RGB images offers a more accessible setting, 
as such images can be readily captured by embodied agents when they navigate an environment.

\begin{figure*}[t]
\centering
\includegraphics[scale=0.54]{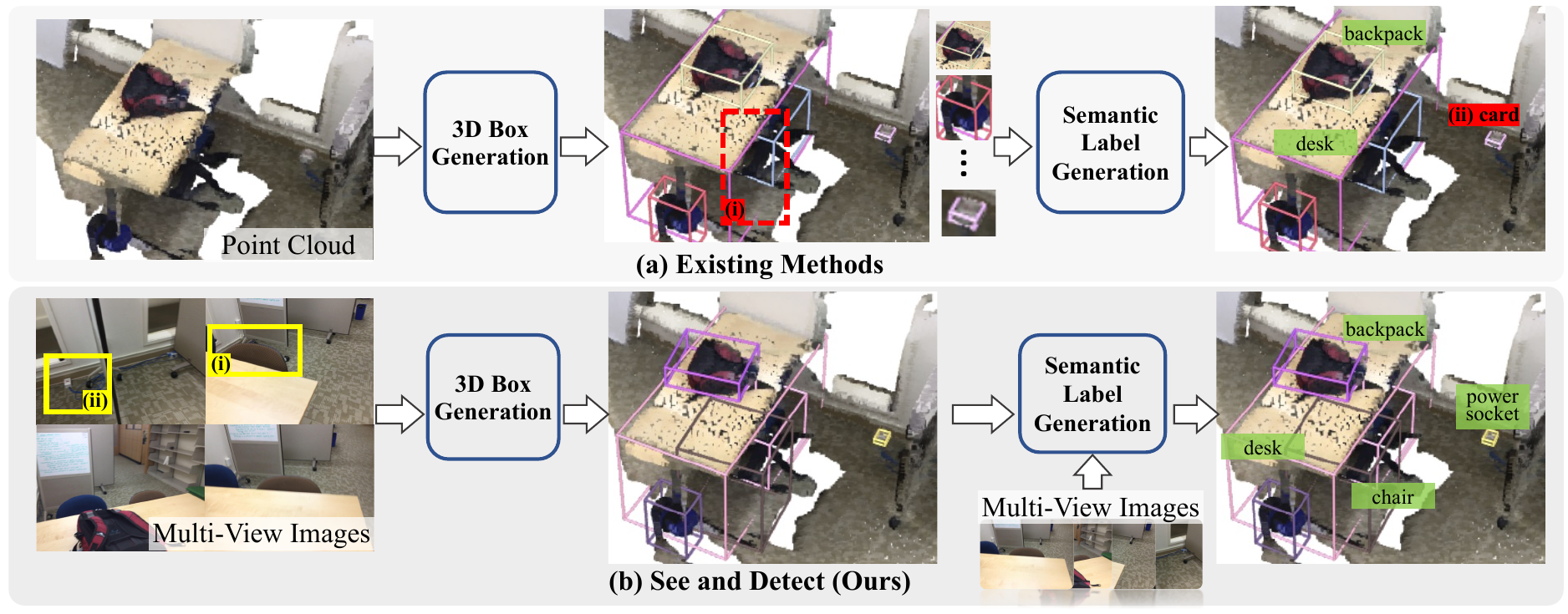}
\caption{Comparison of training pipelines for open-vocabulary multi-view 3D object detection. (a) Existing methods~\cite{lemeshko2025zoo3d, hsu2025openm3d} propose point-cloud-driven approach for 3D box and semantic label generation, which leads to (i) missing object discovery and (ii) incorrect semantic label generation. (b) See and Detect (Ours) proposes a multi-view image-driven approach: objects are discovered through cross-view visual evidence, and semantic labels are generated by considering image contexts. 
}
\label{fig:teaser}
\end{figure*}

Existing multi-view open-vocabulary 3D detection methods, such as OpenM3D~\cite{hsu2025openm3d} and Zoo3D~\cite{lemeshko2025zoo3d}, use posed RGB images as input, but construct training supervision through a point-cloud-driven pipeline.
Specifically, they generate pseudo boxes for class-agnostic localization by lifting 2D image segments into 3D followed by point-cloud validation~\cite{hsu2025openm3d}, or by directly grouping point-cloud segments~\cite{lemeshko2025zoo3d}. The resulting 3D boxes are then projected back to 2D images, where the corresponding image crops are used to infer semantic labels for open-vocabulary classification.
While promising, this design makes both 3D box and semantic label generation mainly determined by point-cloud-derived 3D regions. For box generation, objects with missing or sparse geometry are overlooked, even when they are consistently visible across views (a brown chair that is visible in multiple views is missed by the point-cloud-based method in Fig.~\ref{fig:teaser} (i)). 
For semantic learning, labels inferred from projected regions capture only local appearance and thus ignore the surrounding scene context, leading to wrong predictions for small objects (power socket misclassified to card in Fig.~\ref{fig:teaser} (ii)).
These limitations indicate that existing methods involve images but do not let image evidence drive the key decisions.
In contrast, cross-view visual consistency and scene context can serve as primary evidence for box generation and semantic prediction.



To this end, we propose \textit{See and Detect}, an image-driven framework that uses multi-view images as the primary evidence for both 3D box and semantic label generation.
%
Instead of anchoring pseudo-box generation to reconstructed or sensor-captured point clouds, we derive object proposals from individual RGB views, with depth maps serving as intermediate geometric cues.
Then, to consolidate single-view predictions, we propose
 multi-view consensus box consolidation, where spatially overlapping detections across views are linked into a scene-level graph and nodes with low connectivity are pruned as spurious.
We further propose box quality assessment to reduce the impact of pseudo boxes with partial object coverage by measuring their cross-view visibility consistency and reweighting the localization loss accordingly.

For semantic label generation, we propose context-aware semantic learning to incorporate scene context into semantic label generation. 
Specifically, we query object labels from full images with projected regions highlighted, allowing the prediction to consider global scene context and consequently generating more accurate semantic labels. 
We use an MLLM to perform this context-aware query and average its predictions across views as semantic learning targets.
We further introduce semantic-aware negative sampling for effective contrastive learning, which selects informative negative labels based on semantic similarity in the text feature space. 
Our main contributions are summarized as follows:
\begin{itemize}


\item We develop an image-driven pseudo box supervision pipeline that leverages existing single-view 3D proposals and introduces multi-view consensus box consolidation and box quality assessment to construct reliable localization supervision.

\item We propose context-aware semantic learning, where semantic labels are inferred from full images with projected object regions to incorporate both local appearance and global scene context. We further introduce semantic-aware negative class selection for effective contrastive text--3D alignment.
\item We achieve state-of-the-art performance on both ScanNet and ARKitScenes datasets. Notably, our method outperforms prior work by +4.3\% mAP$_{50}$ on ScanNet60 open-vocabulary detection while running over 500$\times$ faster. 
\end{itemize}
\section{Related Work}
\begin{figure*}[t]
\centering
\includegraphics[scale=0.66]{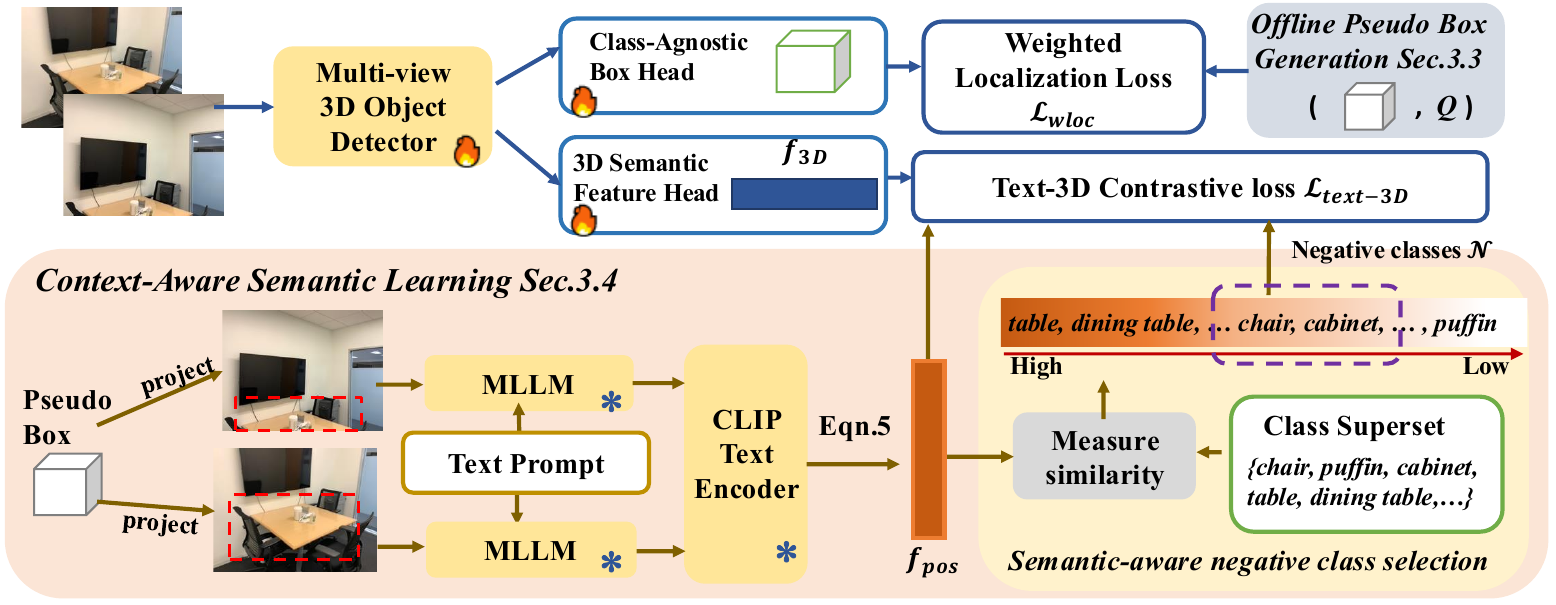}
\caption{Our framework.
During training, it is optimized by two main objectives: (1) a Weighted Localization Loss $\mathcal{L}_{\text{wloc}}$, where generated pseudo 3D boxes are weighted by the quality scores $Q$ (Eqn.~\ref{eqn:quality}) based on their geometric consistency (details shown in Fig.~\ref{fig:box generation}); and (2) a Text-3D Contrastive Loss $\mathcal{L}_{\text{text-3D}}$, which aligns 3D features $f_{3D}$ with semantic labels inferred from MLLM by jointly considering local appearance and global context. We apply Semantic-aware Negative Class Selection to generate informative negative classes.
}
\label{method}
\end{figure*}

\noindent \textbf{Multi-View Indoor 3D Object Detection.}
 Multi-view indoor 3D object detection \cite{shen2024cn, rukhovich2022imvoxelnet, tu2023imgeonet, xu2023nerf, xu2024mvsdet, zhang2025boosting} aims to detect 3D bounding boxes for an indoor scene by taking images as input. Compared to point-cloud-based methods \cite{qi2019deep, rukhovich2023tr3d, misra2021end, zhang2020h3dnet, xie2020mlcvnet, liu2021group} that rely on the expensive depth sensor to capture input data, multi-view based methods are more cost-effective and scalable. 
 The pioneer work, ImVoxelNet \cite{rukhovich2022imvoxelnet}, builds a 3D feature volume in the world space, where each voxel center aggregates corresponding 2D features at its projected pixel location. Then, ImVoxelNet refines the feature volume through a shallow 3D U-Net before predicting 3D bounding boxes from each voxel.
 However, it remains challenging to aggregate correct image features since ground-truth depth is not known.
 Therefore, ImGeoNet \cite{tu2023imgeonet} predicts the surface probability for each voxel to weigh the original feature volume.  It supervises the prediction using ground-truth point cloud.
 Another line of works \cite{xu2023nerf, xu2024mvsdet, zhang2025boosting} propose to predict geometry in a self-supervised way. 
NerfDet \cite{xu2023nerf} queries an opacity score for each voxel from a Neural Radiance Field (NeRF) \cite{mildenhall2021nerf} which is jointly trained with the detection pipeline.
Later, Zhang \etal \cite{zhang2025boosting} reduce the computation of the feature volume by adaptively refining voxels with high occupancy probability.
However, above models remain closed-set, \ie they are restricted to a predefined set of training categories and cannot generalize to unseen classes at inference time.

\noindent \textbf{Open-Vocabulary 3D Object Detection.}
To adapt the 3D object detector in the open world with emerging new classes, open-vocabulary 3D object detection \cite{cao2023coda, cao2025collaborative, lu2023open, jiao2024unlocking, hsu2025openm3d, wang2024ov} is introduced, which can detect new unseen classes in the testing stage given only class names.
The open-vocabulary ability is achieved through class-agnostic object localization and open-vocabulary classification.
OV-3DET \cite{lu2023open} generates class-agnostic boxes by unprojecting the predicted 2D boxes into 3D and refining using point cloud. For the semantic alignment, it introduces De-biased Triplet Cross-Modal Contrastive Learning (DTCC) to align the modality of point cloud, image and text. 
CoDA \cite{cao2023coda} proposes 3D Novel Object Discovery strategy to identify the bounding boxes of novel objects by jointly utilizing the geometry priors from base categories and 2D semantic priors from the CLIP model \cite{radford2021learning}. For semantic alignment, it introduces discovery-driven cross-modal alignment module to align the 3D box features with the corresponding image crop and text label features extracted by CLIP.
However, they all require point cloud as input, which requires depth sensor and is expensive to capture.
Therefore, multi-view open-vocabulary 3D object detection \cite{hsu2025openm3d, lemeshko2025zoo3d} is introduced that does not require point cloud as input. 
Built upon ImGeoNet \cite{tu2023imgeonet}, OpenM3D generates 3D pseudo boxes by lifting and grouping the 2D image segments into 3D and then validating each cluster with point cloud, which are then used for training object localization.
To achieve open-vocabulary classification,
it learns to align the voxel features with the CLIP features of corresponding 2D segments.
Zoo3D \cite{lemeshko2025zoo3d} shares the similar strategy, where pseudo boxes are generated from point cloud instance segmentation and open-vocabulary classification is achieved by comparing the text features with the extracted region-based CLIP features that encode the semantics of the 3D boxes.
Despite their promise, they underutilize images for both box generation and semantic learning.
Using point cloud for checking and aggregating 3D boxes can overlook objects with missing geometry, although observable in the images.
Only using projected image crop to derive semantic label is unreliable as local appearance can be ambiguous and global context is necessary to aid reasoning.
To this end, we propose an image-driven pipeline to fully leverage the multi-view information for effective open-vocabulary 3D detection.


We note that concurrent BoxFusion \cite{lan2025boxfusion} also associates single-view Cubify Anything \cite{lazarow2025cubify} proposals across views. However, it targets training-free online detection, sequentially fusing keyframe proposals with existing detections under causal processing. In contrast, our offline setting enables scene-level association over all proposals, allowing cross-view consensus to suppress isolated predictions (Multi-View Consensus Box Consolidation) and assess imperfect boxes (Box Quality Assessment) for reliable pseudo supervision.

\section{Method}

\begin{figure*}[t]
\centering
\includegraphics[scale=0.7]{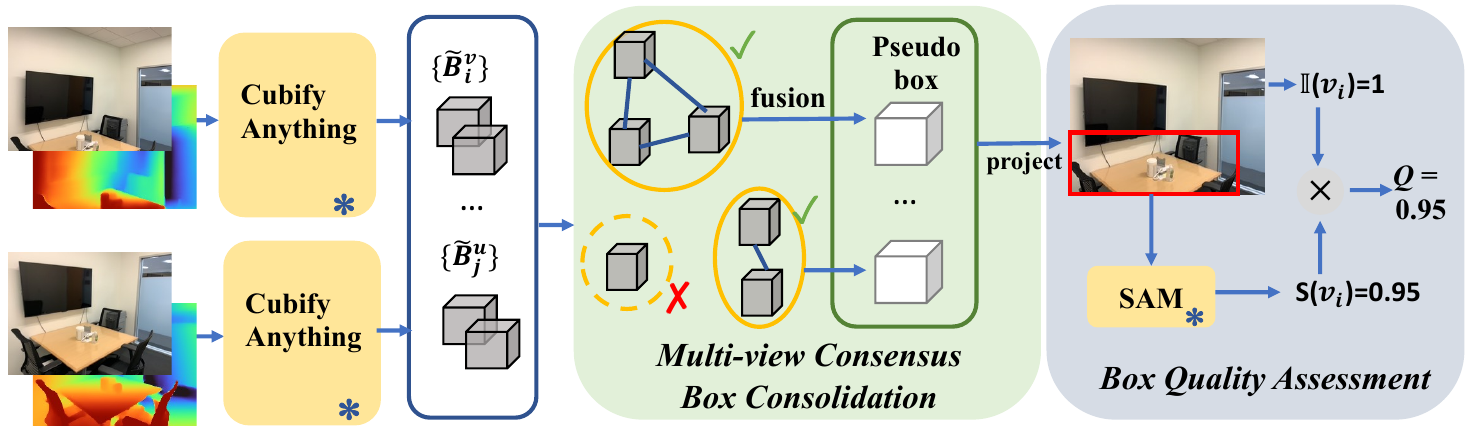}
\vspace{-3mm}
\caption{Image-Driven Pseudo Box Generation. We perform single-view 3D object detection for each image using Cubify Anything. A scene-level undirected graph is constructed where edges represent spatial overlap (Eqn.~\ref{edge}). Multi-View Consensus Box Consolidation is applied to fuse consistent predictions and suppress spurious single-view detections. Finally, each box is assigned a quality score $Q$ through Box Quality Assessment based on cross-view visibility $\mathbb{I}(v_i)$ and SAM-based object coverage check $S(v_i)$, which weights the localization loss during training.
}
\label{fig:box generation}
\vspace{-3mm}
\end{figure*}

In open-vocabulary multi-view indoor 3D object detection (OV-MVI3Det), 
we are given multi-view images of a scene, 
denoted as $\mathcal{I} = \{ I_v \in \mathbb{R}^{3 \times H \times W} \}_{v=1}^{V}$, 
together with the corresponding camera parameters 
$\mathcal{K} = \{ (K_v, H_v)_{v=1}^{V}$\}, 
where $K_v \in \mathbb{R}^{3\times 3}$ denotes the intrinsic matrix, and 
$H_v \in \mathbb{R}^{3 \times 4}$ denotes the extrinsic parameters of the $v$-th view.
The goal of OV-MVI3Det is to localize and classify objects from only multi-view images in the testing stage.

\subsection{Framework Overview}
Fig.~\ref{method} shows our  framework. We adopt SGCDet \cite{zhang2025boosting} as the detection backbone, which is the state-of-the-art multi-view 3D object detection model for indoor scenes.
However, SGCDet is a closed-set detector, where object categories must be predefined during training. To extend it to open-vocabulary 3D detection, we augment the detector with two heads: a class-agnostic box head for object localization and a 3D semantic feature head for open-vocabulary classification.
In the first head, we train for class-agnostic 3D object localization with the pseudo boxes generated via image-driven approach (Sec.~\ref{box generation}), which produces complete and reliable boxes.
In the second head (Sec.~\ref{semantic alignment}), we aim to align the 3D with text feature space, where the context-aware semantic learning is introduced to generate reliable semantic labels by reasoning with scene context. 
We further apply semantic-aware negative class selection for effective text-3D contrastive loss.

\subsection{Image-Driven Pseudo Box Generation}
\label{box generation}

Fig.~\ref{fig:box generation} shows the proposed image-driven pseudo box generation pipeline.
We use Cubify Anything \cite{lazarow2025cubify} as a per-view image-driven proposal generator to predict initial class-agnostic 3D box candidates.
The predicted candidates from the $v$-th view are denoted as
$\{\tilde{B}^v_i\}_{i=1}^{N}=\mathrm{CubifyAnything}(I_v,D_v)$,
where $\tilde{B}^v_i$ is the $i$-th predicted box candidate from image $I_v$, and $D_v$ is the corresponding depth map of $I_v$ to provide 3D information.
These candidates provide visual evidence for object discovery, but are noisy and redundant when generated independently from different views. Therefore, we further perform multi-view consensus box consolidation and box quality assessment to convert the per-view candidates into reliable pseudo 3D boxes for training.

\noindent \textbf{Multi-View Consensus Box Consolidation.}
To associate cross-view predictions, we construct an undirected graph for each scene.
Each node represents a predicted 3D box $\tilde{B}_i^v$.
An edge between $\tilde{B}_i^v$ and $\tilde{B}_j^u$ is established if they have sufficient spatial overlap in 3D:

\begin{equation}
e_{\tilde{B}_i^v,\tilde{B}_j^u} =
\begin{cases}
1, & \text{if } \text{IoU}(\tilde{B}_i^v,\tilde{B}_j^u) > \theta, \\
0, & \text{otherwise},
\end{cases}
\label{edge}
\end{equation}

\noindent where $\text{IoU}(\cdot,\cdot)$ denotes the 3D IoU. 

Then, the resulting graph is decomposed into connected components.
Each component corresponds to a set of cross-view box predictions that agree on a shared 3D location.
We observe that true objects are typically detected consistently across multiple views, forming components with multiple nodes.
In contrast, spurious detections often appear only in a single view, resulting in singleton components.
Therefore, we introduce a multi-view consensus filtering strategy to retain only components containing more than one node and discard singleton components.
For each retained component, we fuse all boxes within the component to generate a consolidated pseudo 3D box. 
Importantly, during fusion, we also record the associated view indices of all boxes within the component. 
For the $k$-th component, we denote its supporting view set as $\mathcal{V}_k = \{ v_{i} \}$, where $v_{i}$ is the view from which a box in the component is predicted. 


\begin{figure*}[t]
\centering
\includegraphics[scale=0.73]{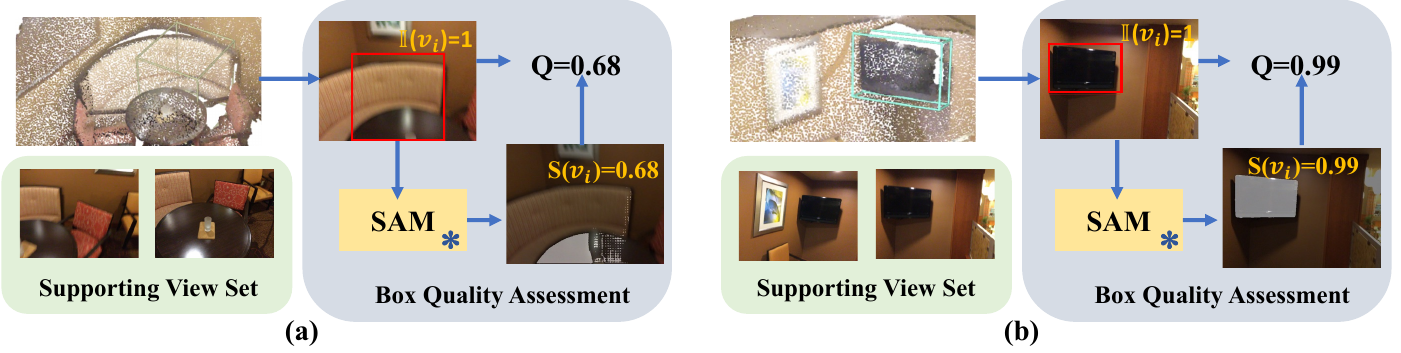}
\vspace{-1mm}
\caption{Box quality assessment. (a) shows a pseudo box with incomplete object coverage due to detection from local image views. (b) shows a pseudo box with complete object coverage. We project the boxes to the nearby view $v_i$ for box quality assessment. The quality score $Q$ can correctly reflect box quality. 
}
\label{fig:box quality}
\vspace{-3mm}
\end{figure*}

\noindent \textbf{Box Quality Assessment.}
However, the fused boxes may correspond to incomplete object coverage, especially when they are formed from limited local views.
As illustrated in Fig.~\ref{fig:box quality} (a), only a partial sofa is detected since all the supporting views show part of the sofa.
To reduce the influence of inferior pseudo boxes during training, we propose box quality assessment.
Our key observation is that a high-quality 3D box should satisfy two geometric consistency properties: (1) it should remain visible when projected to nearby views; (2) its projection should tightly cover the object instance rather than only a partial region.
Therefore, we evaluate each pseudo box by checking visibility and object coverage in the neighbouring views of its supporting view set.

For each pseudo box, we select the view with the highest prediction confidence as the anchor view,
and project the box to its nearby views denoted as $v_i$. 
Let $\bar{P} \in \mathbb{R}^{8 \times 4}$ denote the homogeneous coordinates of the 3D box corners.
The projection to view $v_i$ is computed as $\bar{p}^\top
=
K_{v_i}
H_{v_i}
\bar{P}^\top$,
followed by perspective division to obtain 2D projected corners
$p = [p_1,\dots, p_8]^\top \in \mathbb{R}^{8\times 2}$,
where $p_j = (u_j, v_j)$.
A projected corner is considered visible if it lies within the image boundary,
\ie, $0 < u_j < W$ and $0 < v_j < H$.
We define the visibility indicator for view $v_i$ as
\begin{equation}
\mathbb{I}(v_i)
=
\mathbf{1}
\left\{
\sum_{j=1}^{8}
\mathbf{1}(0 < u_j < W,\; 0 < v_j < H)
\geq 4
\right\}.
\end{equation}

For visible projection, we further check object coverage using SAM \cite{kirillov2023segment}.
Our intuition is that a projection that tightly covers the instance is a good prompt for SAM, leading to high-confidence mask prediction. 
Therefore, we use the minimum and maximum coordinates of the projected corners $p$ as the prompt.
We denote the SAM confidence as $S(v_i) \in [0,1]$. The final box quality score is computed as
\begin{equation}
Q
=
\frac{1}{n}
\sum_{i=1}^{n}
\mathbb{I}(v_i)\, S(v_i),
\label{eqn:quality}
\end{equation}
where $n$ is the number of nearby views and is empirically set to 3. 
During training, $Q$ is used as the weight for object localization loss so that imperfect pseudo boxes contribute less to optimization.

\subsection{Context-Aware Semantic Learning}
\label{semantic alignment}
To leverage the rich context information in the images, we use a multi-modal large language model (MLLM) to infer semantic label for each pseudo box. 
Specifically, we project each pseudo box onto its supporting views and overlay the projected 2D box on the image as a visual prompt for Qwen3-VL \cite{bai2025qwen3}, enabling category prediction with both local object appearance and scene context.
Let $\hat{y}_i$ denote the predicted label from the $i$-th supporting view, and 
$f_{\text{text}}^{i} = E_{\text{text}}(\hat{y}_i) \in \mathbb{R}^{d}$ is the extracted text feature by CLIP.
The semantic label for the pseudo box is computed as follows:
\begin{equation}
f_{\text{pos}} = \frac{1}{n_{\text{supp}}}\sum_{i=1}^{n_{\text{supp}}} f_{\text{text}}^{i},
\label{eq:pos_proto}
\end{equation}
where $n_{\text{supp}}$ is the number of supporting views of the pseudo box.

\noindent \textbf{Semantic-Aware Negative Class Selection.}
To perform effective text-3D contrastive learning, the quality of negative classes matters. Therefore,
we construct semantic-aware negative classes by measuring the cosine similarity between the semantic label $f_{\text{pos}}$ and the text features of a class superset $\mathcal{T}=\{t_m\}_{m=1}^{M}$. Classes are then sorted in descending order based on the similarity scores.
Classes ranked at the top are highly similar and may correspond to
the true class or near-duplicate categories,
while classes ranked at the bottom are semantically unrelated
and provide limited discriminative learning signal.
Therefore, we select the negative classes from a middle range denoted as $\mathcal{N}=\{t_{\pi(k)} \mid k_1 \le k \le k_2\}$, where $\pi(k)$ represents the index of class ranked at position $k$.

\noindent \textbf{Optimization.}
To enable open-vocabulary classification, we convert the classification head of SGCDet to predict a fixed-length feature vector that is aligned with the CLIP feature space. 
We denote the feature vector as $f_{\text{3D}} \in \mathbb{R}^d$. The text-3D contrastive loss is as follows:
\begin{equation}
\mathcal{L}_{\text{text-3D}}
=
-\log
\frac{
\exp\!\left(\cos(f_{3D}, f_{\text{pos}})/\tau\right)
}{
\sum\limits_{f \in \mathcal{M}}
\exp\!\left(\cos(f_{3D}, f)/\tau\right)
},
\label{text-3D loss}
\end{equation}
where $\mathcal{M}=\{f_{\text{pos}} \cup \{ E_{\text{text}}(t) \mid t\in\mathcal{N}\} \}$. $\tau$ is the temperature hyper-parameter and set to 0.1. 
The network is jointly trained with object localization loss \cite{zhang2025boosting} weighted by our proposed box quality assessment score $\mathcal{L}_{\text{wloc}}$ and semantic learning objective  $\mathcal{L}_{\text{text-3D}}$  to achieve open-vocabulary 3D object detection.

\section{Experiment}

\subsection{Datasets}
We evaluate on two challenging 3D indoor detection datasets: ScanNet \cite{dai2017scannet} and ARKitScenes \cite{baruch2021arkitscenes}.
ScanNet has 1,201 training and 312 testing scenes with annotations for 200 object categories. 
We evaluate on two benchmarks, \ie ScanNet60 with 60 classes \cite{cao2023coda} and ScanNet200 with 200 classes \cite{hsu2025openm3d}. 
We report mean average precision (mAP) under IoU thresholds of 0.25 and 0.5.
ARKitScenes consists of 4493 training and 549 testing scans. These scans contain RGB-D frames along with 3D object bounding box annotations of 17 categories. 
Following \cite{hsu2025openm3d}, we use ARKitScenes only for 3D pseudo box evaluation and report precision and recall under IoU thresholds of 0.25 and 0.5.

\subsection{Implementation Details}
Our detection backbone is SGCDet \cite{zhang2025boosting}, which is the state-of-the-art multi-view indoor 3D object detector. 
We follow the training and testing configuration of SGCDet. 
Images are resized to $320 \times 240$ as the input with 40 and 100 images for training and testing, respectively.
The network is trained end-to-end under 50 epochs using AdamW optimizer.
In the pseudo box generation, we evenly sample 60 views for each scene to detect bounding boxes. We perform Non-Maximum Suppression and size constraints after multi-view consensus box consolidation. $\theta$ is set to 0.4 in the edge construction. 
In the context-aware semantic learning, we follow CoDA \cite{cao2023coda} to adopt LVIS \cite{gupta2019lvis} categories as the class superset. $k_1$ and $k_2$ are set to 10 and 150 in the semantic-aware negative class selection. 
We use Qwen3-VL-30B-A3B-Instruct \cite{bai2025qwen3} as the MLLM for semantic label generation.
During testing stage, we measure the cosine similarity between the predicted 3D box feature $f_{\text{3D}}$ and CLIP text embeddings of text prompts to perform open-vocabulary classification.
Following CoDA \cite{cao2023coda}, the template for text prompt is set as ``A photo of [category name] in the scene''.
Importantly, Cubify Anything, SAM, and Qwen3-VL are used only for offline pseudo-label generation during training. At inference time, our method uses only the feed-forward multi-view 3D detector with RGB images and camera parameters as input.

\subsection{Comparison with Baselines}

\begin{figure*}[t]
\centering
\includegraphics[scale=0.6]{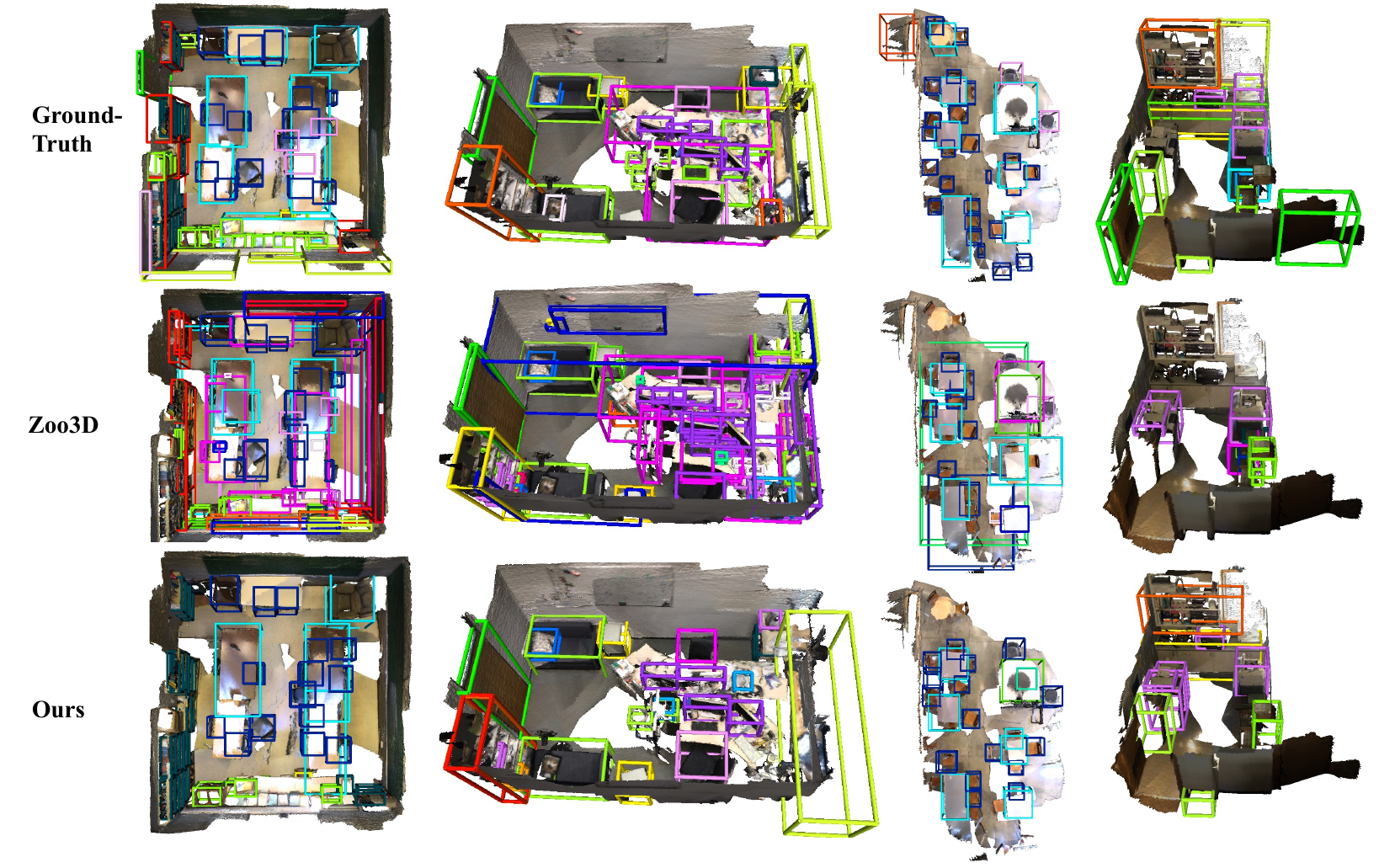}
\vspace{-3mm}
\caption{Qualitative results of open-vocabulary multi-view 3D object detection on ScanNet60.
}
\label{fig:scannet60}
\vspace{-1mm}
\end{figure*}

\noindent \textbf{Open-Vocabulary 3D Object Detection.}
Tab.~\ref{ov-scannet} compares our method with prior open-vocabulary 3D object detection approaches on ScanNet60 and ScanNet200 benchmarks.
The baselines include state-of-the-art point-cloud-based methods such as CoDA \cite{cao2023coda}, OV-3DET \cite{lu2023open} and OV-Uni3DETR \cite{wang2024ov}, as well as multi-view-based approaches including OpenM3D \cite{hsu2025openm3d} and Zoo3D \cite{lemeshko2025zoo3d}.
We report the best results in the setting of posed images from Zoo3D \cite{lemeshko2025zoo3d}, which learns a class-agnostic 3D object detection from the pseudo boxes generated from point cloud instance segmentation and then extracts the CLIP features of the projected 2D box crops for open-vocabulary classification.
Despite taking only RGB images as input, our method largely outperforms point-cloud-based open-vocabulary detectors (\eg improves over OV-Uni3DETR by +9.1 on ScanNet60).
Compared with multi-view based methods, our approach also achieves state-of-the-art performance. 
In particular, we outperform Zoo3D by +4.6 mAP$_{25}$ and +4.3 mAP$_{50}$ on ScanNet60.
It verifies the superiority of the proposed image-driven approach for pseudo box and semantic label generation, which provides more complete 3D boxes and more accurate semantic labels than point-cloud-driven approaches (\ie Zoo3D and OpenM3D).
Moreover, we run much faster than Zoo3D during inference, achieving a 500x speedup.
The large speed gap stems from the fact that Zoo3D performs online 3D reconstruction and aligns 3D boxes to 2D image crops for open-vocabulary classification at inference time, while our method is a single-stage feed-forward 3D detector.
Fig.~\ref{fig:scannet60} shows qualitative comparison on ScanNet60, where we show more compact and accurate open-vocabulary 3D object detections.

\begin{table*}[t]
\centering
\resizebox{\linewidth}{!}{
\small
\begin{tabular}{p{3.5cm}<{\centering} p{1.5cm}<{\centering} p{1.5cm}<{\centering} p{1.5cm}<{\centering} p{1.5cm}<{\centering} p{1.5cm}<{\centering} p{2.5cm}<{\centering}}
\toprule
\multirow{2}{*}{Method} & \multirow{2}{*}{Test Input} & \multicolumn{2}{c}{ScanNet60} & \multicolumn{2}{c}{ScanNet200} & \multirow{2}{*}{Inference Time(s)} \\
                        &                                 & mAP$_{25}$          & mAP$_{50}$         & mAP$_{25}$          & mAP$_{50}$          &                                                                            \\ \hline
CoDA \cite{cao2023coda}                          & pcd        & 9.0             & 2.2          & 4.4            & 0.9             & 1.8      \\
OV-3DET \cite{lu2023open}                           & pcd       & 8.1               & 1.4             &2.7                & 0.5             & 2.6      \\
OV-Uni3DETR \cite{wang2024ov}                     & pcd        & 19.4               &  -             &  -              & -               & -    \\
\midrule
OpenM3D \cite{hsu2025openm3d}             & imgs     & -                &  -             & 4.2               & -               & 0.4   \\ 
Zoo3D \cite{lemeshko2025zoo3d}      & imgs     & 23.9               &  10.8            & \textbf{16.5}              & 6.3              & 378.3  \\
Ours        & imgs    & \textbf{28.5}               &  \textbf{15.1}            & 16.4              & \textbf{7.7}              & 0.7  \\
\bottomrule
\end{tabular}
}
\vspace{-3mm}
\caption{Open-vocabulary multi-view 3D object detection on ScanNet60 and ScanNet200 benchmarks. 
``Test Input'' denotes the input at inference stage. 
Inference time is measured in seconds per scene. 
We report the best results of Zoo3D from their paper under the multi-view setting. Best results are marked bold.}
\label{ov-scannet}
\vspace{-3mm}
\end{table*}

\begin{table}
\small
\centering
\resizebox{\linewidth}{!}{
\begin{tabular}{ccccc}
\toprule
Method   & mAP$_{25}$          & mAP$_{50}$         & mAR$_{25}$          & mAR$_{50}$  \\ 
\midrule
OpenM3D \cite{hsu2025openm3d} & 26.9   & -   & 51.2   & -  \\ 
Zoo3D \cite{lemeshko2025zoo3d}  & 36.1   & 13.9      & -     & -     \\ 
Ours        & \textbf{52.7}     & \textbf{25.3}    & \textbf{72.2}     & \textbf{38.9}     \\ \bottomrule
\end{tabular}
}
\vspace{-3mm}
\caption{Class-agnostic 3D object detection on ScanNet200. Best results are marked bold.}
\label{class-agnostic det}
\vspace{-5mm}
\end{table}

\noindent \textbf{Class-Agnostic 3D Object Detection.}
Tab.~\ref{class-agnostic det} reports the class-agnostic 3D detection performance on ScanNet200.
In this setting, the detector predicts foreground 3D bounding boxes without category supervision, thus directly evaluating the quality of pseudo boxes.
Our method significantly outperforms prior approaches across all metrics. 
In particular, we achieve 52.7 mAP$_{25}$ and 25.3 mAP$_{50}$,
surpassing Zoo3D by +16.6 at mAP$_25$ and +11.4 at mAP$_50$, respectively.
Compared to OpenM3D, our method improves mAP$_{25}$ by +25.8 (26.9 $\rightarrow$ 52.7).
It verifies the superiority of the proposed image-driven pseudo box generation pipeline, which provides high-quality box training data for class-agnostic localization.

\noindent \textbf{3D Pseudo Box Evaluation.}
Tab.~\ref{pseudo box} shows the evaluation of 3D pseudo boxes on ScanNet200 and ARKitScenes datasets. 
We achieve best recall and precision, which verifies the proposed image-driven pseudo box generation method with multi-view consensus box consolidation provides complete and accurate pseudo boxes.


\begin{table*}[t]
\centering
\small
\resizebox{\linewidth}{!}{
\begin{tabular}{p{2.cm}<\centering   p{1.2cm}<\centering p{1.2cm}<\centering p{1.2cm}<\centering p{1.2cm}<\centering p{1.2cm}<\centering p{1.2cm}<\centering p{1.2cm}<\centering p{1.2cm}<\centering}
\toprule
\multirow{2}{*}{Method}  & \multicolumn{4}{c}{ScanNet200} & \multicolumn{4}{c}{ARKitScenes} \\ \cline{2-9} 
                          & R$_{25}$    & P$_{25}$    & R$_{50}$   & P$_{50}$   & R$_{25}$    & P$_{25}$    & R$_{50}$   & P$_{50}$   \\ \midrule
OpenM3D \cite{hsu2025openm3d}      & 58.3     & 32.1    & 32.9    & 18.1    & 51.9     & 5.9     & 13.7    & 1.6    \\
Zoo3D \cite{lemeshko2025zoo3d}          & 58.7     & 35.9    & 45.6    & 27.9    &  52.0    & 8.4     & 28.2    & 4.3      \\
Ours          & \textbf{71.9}    & \textbf{50.5}    & \textbf{59.6}    & \textbf{41.9}    &  \textbf{61.3}    & \textbf{45.6}     & \textbf{58.0}    & \textbf{43.2}      \\ 
\bottomrule
\end{tabular}
}
\vspace{-3mm}
\caption{3D pseudo box evaluation on ScanNet200 and ARKitScenes dataset. We report recall (R) and precision (P) at IoU thresholds of 0.25 and 0.50. Best results are marked bold.}
\label{pseudo box}
\end{table*}

\subsection{Ablation Study}

\begin{table*}[t]
\begin{minipage}{0.46\linewidth}
\centering
\resizebox{\linewidth}{!}{
\begin{tabular}{ccccc}
\toprule
Method               & R$_{25}$  & P$_{25}$  & R$_{50}$  & P$_{50}$  \\ \midrule
No Filtering            & \textbf{76.5} & 34.1 & \textbf{61.2} & 27.3 \\
Filtering ($<3$ views)   & 65.5   & \textbf{53.7}   & 56.3   & \textbf{43.9}   \\
Ours                 & 71.9 & 50.5 & 59.6 & 41.9 \\ 
\bottomrule
\end{tabular}
}
\vspace{-3mm}
\caption{Ablation on multi-view consensus box consolidation on ScanNet200. ``No Filtering''  and 
``Filtering ($< 3$ views)'' denote no filtering and 
filtering components observed less than 3 views, respectively.
}
\label{multi-view filtering}
\end{minipage}
\hspace{0.2cm}
\begin{minipage}{0.5\linewidth}
\centering
\resizebox{\linewidth}{!}{
\begin{tabular}{ccccc}
\toprule
Method            & mAP$_{25}$ & mAP$_{50}$ & mAR$_{25}$ & mAR$_{50}$ \\ \midrule
No Asses.         & 50.5  & 23.6  & 71.1  & 38.6  \\
Visibility Asses. & 51.9   & 24.1  & 71.5    & 38.9    \\
Ours              & \textbf{52.7}  & \textbf{25.3}  & \textbf{72.2}  & \textbf{38.9}  \\ \bottomrule
\end{tabular}
}
\vspace{-3mm}
\caption{Ablation study on the box quality assessment on ScanNet200. 
``No Asses.'' denotes no box quality assessment, while
``Visibility Asses.'' denotes only checking visibility.}
\label{quality assessment}
\end{minipage}
\end{table*}

\noindent \textbf{Effectiveness of Multi-View Consensus Box Consolidation.}
Tab.~\ref{multi-view filtering} shows the ablation study of multi-view consensus filtering during pseudo box generation on ScanNet200. ``No Filtering'' denotes no multi-view consensus filtering, \ie keep all components.
``Filtering ($< 3$ views)'' refers to removing components that have fewer than 3 nodes (\ie observed in less than 3 views).
``No Filtering'' shows the highest recall (76.5 / 61.2 at R$_{25}$/R$_{50}$) but very low precision (34.1 / 27.3), as many spurious single-view detections are preserved. 
By removing only singleton components, precision improves significantly (e.g., P$_{50}$ increases from 27.3 to 41.9) while recall at IoU 0.5 drops only slightly (61.2 $\rightarrow$ 59.6), indicating that most false positives originate from single-view predictions and can be effectively suppressed without sacrificing valid objects. 
Further applying stricter filtering (``Filtering ($< 3$ views)'') leads to very marginal improvement on precision but at a sacrifice of substantial decrease in recall.
Based on this precision--recall trade-off, we adopt the strategy of removing only singleton components in the multi-view consensus filtering.

\noindent \textbf{Effectiveness of Box Quality Assessment.}
Tab.~\ref{quality assessment} shows the ablation study on the box quality assessment on the class-agnostic 3D object detection on ScanNet200.
``No Asses.'' denotes no box quality assessment and $Q=1$ for all boxes in the training. 
``Visibility Asses.'' denotes only checking visibility $\mathbb{I}(v_i)$ to compute $Q$.
``No Asses.'' and
``Visibility Asses.'' are worse than ours because they do not fully assess pseudo box quality.

\begin{table}[t]
\centering
\resizebox{\linewidth}{!}{
\begin{tabular}{ccccc}
\toprule
\multirow{2}{*}{Method} & \multicolumn{2}{c}{ScanNet60} & \multicolumn{2}{c}{ScanNet200} \\
                        & mAP$_{25}$          & mAP$_{50}$         & mAP$_{25}$          & mAP$_{50}$         \\ \midrule
IMG Crop                    & 20.6             & 11.4       & 7.8         & 4.4            \\ 
IMG Crop+Text            &  21.0              &  11.7       & 9.9         & 4.5              \\ 
Ours & \textbf{28.5} & \textbf{15.1} & \textbf{16.4} & \textbf{7.7} \\
\bottomrule
\end{tabular}
}
\vspace{-3mm}
\caption{Ablation study on context-aware semantic learning. ``IMG Crop`` and ``IMG Crop+Text'' are two semantic learning methods derived from image crops.}
\label{mllm-label}
\end{table}

\begin{table}[t]
\centering
\resizebox{\linewidth}{!}{
\begin{tabular}{ccccc}
\toprule
\multirow{2}{*}{Range} & \multicolumn{2}{c}{ScanNet60} & \multicolumn{2}{c}{ScanNet200} \\
                        & mAP$_{25}$          & mAP$_{50}$         & mAP$_{25}$          & mAP$_{50}$         \\ \midrule
$[2,1203]$        &  27.0       &  13.4      & 15.0         & 6.8            \\ 
$[150,400]$       &  28.1       &  14.0      & 15.5         & 7.0              \\ 
$[400,800]$      &  27.1         & 13.0       & 15.0         & 6.5              \\
Ours & \textbf{28.5} & \textbf{15.1} & \textbf{16.4} & \textbf{7.7} \\
\bottomrule
\end{tabular}
}
\vspace{-3mm}
\caption{Ablation study on the range $[k_1, k_2]$ for the semantic-aware negative class selection on ScanNet60 and ScanNet200.}
\label{negative-class}
\vspace{-3mm}
\end{table}

\noindent \textbf{Effectiveness of Context-Aware Semantic Learning.}
Tab.~\ref{mllm-label} shows the ablation study on context-aware semantic learning.
We compare with the semantic alignment strategies that rely on image crops.
``IMG Crop'' is the class-agnostic distillation loss \cite{cao2023coda},
which aligns 3D with image region features.
``IMG Crop + Text'' further incorporates the DTCC loss~\cite{lu2023open},
which introduces additional text supervision derived from region-level CLIP similarities.
However, both models show worse performance than ours, indicating that semantic alignment built upon local image crops alone is limited.
In contrast, we leverage MLLM to jointly reason the global scene context and local object region, achieving effective semantic alignment.

\noindent \textbf{Analysis of selection range for negative class.}
Tab.~\ref{negative-class} shows the experiments of negative class selection on the open-vocabulary detection on ScanNet60 and ScanNet200 benchmarks.
Using the full range $[2,1203]$ (excludes the top-1 most similar class) leads to sub-optimal performance.
This is because the selected negatives include near-duplicative categories semantically too close to the positive class and many trivial easy negatives far away in the embedding space, consequently diluting informative negatives.
We then evaluate different middle ranges and observe that selecting negatives from the upper-middle band $[10,150]$ yields the best performance.

\section{Conclusion}
In this paper, we present See and Detect, an image-driven framework for open-vocabulary multi-view indoor 3D object detection. Unlike prior point-cloud-driven pipelines, our method treats multi-view images as the primary evidence for both class-agnostic localization and open-vocabulary classification.
For localization, we leverage existing single-view 3D proposals and introduce multi-view consensus box consolidation to suppress spurious predictions, together with box quality assessment to down-weight imperfect boxes.
For semantic learning, we introduce context-aware semantic learning, where semantic labels are inferred from full images overlaid with projected object boxes, allowing both local object appearance and global scene context to be effectively exploited. In addition, semantic-aware negative class selection is designed to improve contrastive text--3D alignment. 
We achieve state-of-the-art performance on ScanNet and ARKitScenes.
\clearpage
{
    \small
    \bibliographystyle{ieeenat_fullname}
    \bibliography{main}
}

\end{document}